\documentclass[conference]{IEEEtran}

\usepackage{times}
\usepackage{latexsym}
\usepackage{amsmath}
\usepackage{multirow}
\usepackage{url}
\usepackage{graphicx}
\usepackage{caption}
\usepackage{subcaption}
\usepackage{tabularx}
\usepackage{url}
\usepackage{amsmath}
\usepackage{multirow}
\usepackage{CJKutf8}
\newcolumntype{Y}{>{\centering\arraybackslash}X}

\ifCLASSINFOpdf
\else
\fi

\begin{document}

\title{Aspect Sentiment Model for Micro Reviews}


\author{\IEEEauthorblockN{Reinald Kim Amplayo}
\IEEEauthorblockA{
Yonsei University\\
Seoul, South Korea\\
Email: rktamplayo@yonsei.ac.kr}
\and
\IEEEauthorblockN{Seung-won Hwang}
\IEEEauthorblockA{
Yonsei University\\
Seoul, South Korea\\
Email: seungwonh@yonsei.ac.kr}}

\maketitle

\begin{abstract}
 This paper aims at an aspect sentiment model for aspect-based sentiment analysis (ABSA) focused on micro reviews. This task is important in order to understand short reviews majority of the users write, while existing topic models are targeted for expert-level long reviews with sufficient co-occurrence patterns to observe. Current methods on aggregating micro reviews using metadata information may not be effective as well due to metadata absence, topical heterogeneity, and cold start problems. To this end, we propose a model called Micro Aspect Sentiment Model (MicroASM). MicroASM is based on the observation that short reviews 1) are viewed with sentiment-aspect word pairs as building blocks of information, and 2) can be clustered into larger reviews. When compared to the current state-of-the-art aspect sentiment models, experiments show that our model provides better performance on aspect-level tasks such as aspect term extraction and document-level tasks such as sentiment classification. 
\end{abstract}

\IEEEpeerreviewmaketitle

\section{Introduction}

Recently, the number of short reviews is increasing as most people write concise and brief reviews, while only few enthusiasts write long and expertly-written reviews. Following this phenomenon, some websites \footnote{Naver Movies (URL: \url{http://movie.naver.com/}) and Douban Movies (URL: \url{https://movie.douban.com/})} are already implementing a rule forcing users to limit their reviews to a specified number of characters, usually the same as a tweet with 140 characters.
With the increase of short reviews, it is necessary to create models specific to them.


Our research goal is thus, to holistically understand the opinions of the general people, yet in the quality of expert reviews, by performing an aspect-based sentiment analysis (ABSA) on short reviews. Specifically, we pursue LDA-based approaches \cite{blei2003latent}, which are previously used as a semi-supervised technique to answer several ABSA tasks \cite{lin2009joint,jo2011aspect,wang2016mining}.
LDA-based approaches are advantageous over other approaches since they only require a small sentiment lexicon and do not require labelled training data, unlike other recent deep learning-techniques \cite{wang2016recursive,wang2016attention,ruder2016hierarchical}. However, conventional topic models are not fit, as they are designed to be used on documents that are long enough to extract meaningful discoveries. Empirical studies also show that these models do not perform well on short texts when experimented on classification and prediction problems \cite{hong2010empirical}. This is because, when fed with short texts, topic models suffer from data sparsity due to the lack of co-occurrence patterns \cite{yan2013biterm}.

Current solutions are proposed to remedy the problem of aspect sentiment models by a simple aggregation or pooling of short documents, based on meta-information available on documents such as time slices \cite{kaminka2016joint} and hashtags \cite{ahuja2016microblog}. However, there are three major problems in current methods. First, document metadata such as users and hashtags are not always available. Reviews seldom use hashtags and there are lots of anonymous users. Second, documents aggregated using these approaches tend to be topically heterogeneous \cite{alvarez2016topic}. This is because the assumptions made by aggregation techniques frequently violate the topic consistencies assumed by the model. Lastly, there are cold start entities without enough documents to be pooled into one large document. This makes the documents of these users less informative and may be disregarded by the topic model.

Different from normal-sized reviews, short reviews have several properties. We observe two properties on short reviews. First, we observe that short reviews consist of pairs of sentiment word and aspect word. We call these pairs \textbf{sentiment-aspect pairs}. For example, in a review \textit{``Unprofessional service and indifferent staff. Disappointed because of nice ambience.''}, we can extract sentiment-aspect pairs \textit{unprofessional service}, \textit{indifferent staff}, \textit{disappointed staff}, and \textit{nice ambience}\footnote{We note that a sentiment word is not restricted to adjectives as shown in the examples; this includes opinionated verbs and negated words as well.}. Instead of generating words a word at a time, modeling sentiment-aspect pairs results to a \textbf{higher coherence}. This means that words in sentiment-aspect pairs will appear with much higher probabilities in the same aspect distribution. This improves the performance of the model on aspect-level tasks.

Second, we observe that short reviews can be grouped into clusters based on their aspect distributions. One can think of the clusters as large and unorganized list of words, which when properly organized, can be reconstructed into expertly-written reviews. For example, the sentences \textit{``Despite the bad food, I liked the atmosphere and the kind waiters.''} and \textit{``The pasta was really bad. They treated us like kings, though.''} can be grouped together into one cluster because of their similar positive sentiments on atmosphere and service, and negative sentiment on food. This may result into having the reviews in the cluster \textbf{share information}. This improves the detection of aspect and sentiment proportions in the document, which in turn improves the performance of the model on document-level tasks.

\section{Micro Aspect Sentiment Model}

\begin{table}[t]
    \small
    \centering
    \caption{Meanings of the notations in MicroASM}
    \label{tab:notations}
    \begin{tabularx}{0.45\textwidth}{|cX|}
        \hline
        $D$ & \# of documents \\
        $C$ & \# of clusters \\
        $N$ & \# of words in a document \\
        $P$ & \# of pairs in a document \\
        $S$ & \# of sentiments \\
        $T$ & \# of topics \\
        $c$ & cluster \\
        $w$ & word \\
        $s$ & sentiment \\
        $z$ & aspect \\
        $\theta$ & aspect multinomial distribution of a cluster \\
        $\phi$ & word multinomial distribution of an aspect-sentiment pair \\
        $\pi$ & sentiment multinomial distribution of a cluster \\
        $\psi$ & cluster multinomial distribution \\
        $\alpha$ & Dirichlet prior for $\theta$ \\
        $\beta$ & Dirichlet prior for $\phi$ \\
        $\gamma$ & Dirichlet prior for $\pi$ \\
        $\delta$ & Dirichlet prior for $\psi$ \\
        \hline
    \end{tabularx}
\end{table}

\begin{figure}[t]
    \centering
    \includegraphics[height=2.3in]{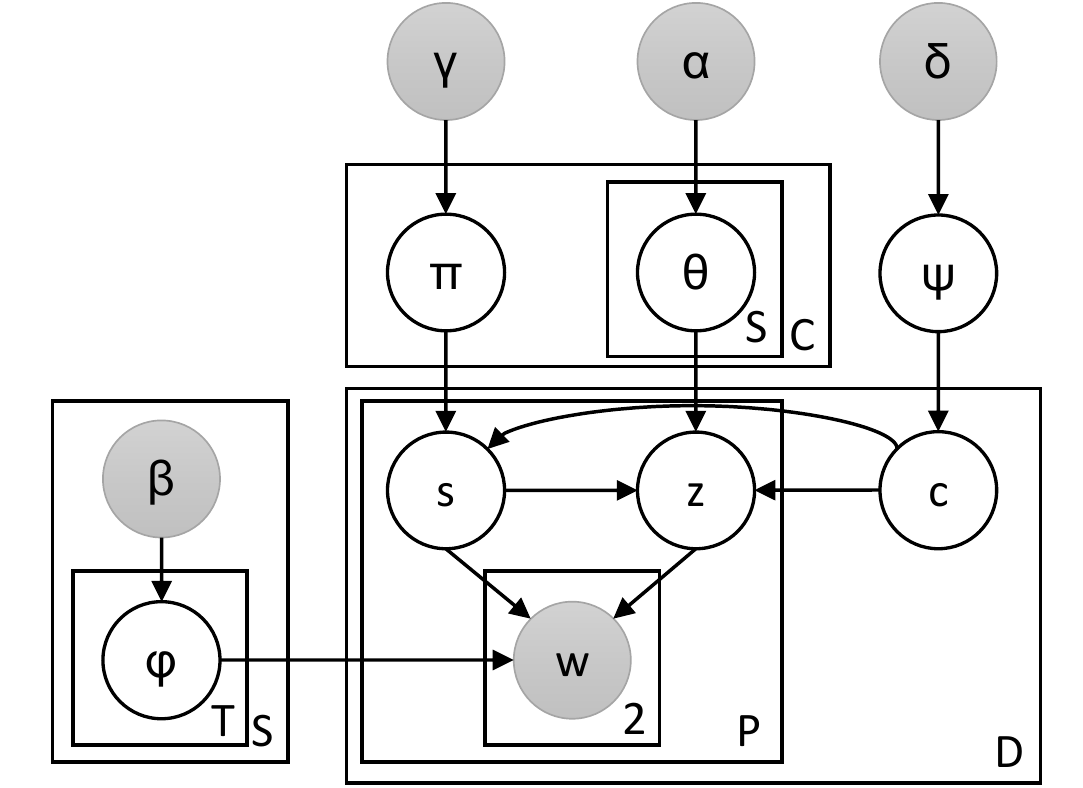}
    \caption{Graphical representation of MicroASM}
    \label{fig:pcasm}
\end{figure}

Based on these observations, we present Micro Aspect Sentiment Model (MicroASM). We provide the source code of the model in the link for reproducibility: \url{https://github.com/rktamplayo/MicroASM}. The model is shown graphically in Figure \ref{fig:pcasm}. Notations are summarized in Table \ref{tab:notations}. Instead of generating just one word at a time, MicroASM generates pairs of words using both the aspect and the sentiment latent variables $z$ and $s$. Note that we include not only sentiment-aspect pairs but also all kinds of word pairs (i.e. sentiment-sentiment pairs and aspect-aspect pairs among others). This addition helps in two ways. First, it removes the necessity of doing aspect word and sentiment word detection beforehand. Second, modeling sentiment-sentiment and aspect-aspect pairs enhances sentiment and aspect learning, respectively. Moreover, outside the document layer, MicroASM introduces a new cluster layer. The cluster layer holds the sentiment and aspect distributions $\pi$ and $\theta$, respectively. In order to connect both document and cluster layer, MicroASM introduces a new cluster latent variable $c$ inside the document layer. Pairs of words are then generated using both $z$ and $s$, both of which depend on $c$. Following the traditional aspect sentiment models \cite{lin2009joint,jo2011aspect}, we use prior sentiment lexicons by encoding it to the Dirichlet prior $\beta$. This leads to an assymetric $\beta$ vector that controls the sentiment of the word distribution.

\subsection{Generative process}

In MicroASM, pairs of words are generated using both the aspect and the sentiment latent variables $z$ and $s$ respectively. Both aspect and sentiment variables are dependent to the cluster variable $c$. The probability of a pair of words $w_1$ and $w_2$ in document $d$ is given in Equation \ref{eq:pcasmgen1}. Using the ideas on factorization and parameterization described above, we do the same as shown in Equation \ref{eq:pcasmgen2}.
\begin{align}
    &P(w_1,w_2) \nonumber \\
    \label{eq:pcasmgen1}
    & = \sum_c \sum_s \sum_z P(c|d) P(s,z|c) P(w_1|s,z) P(w_2|s,z) \\
    \label{eq:pcasmgen2}
    & = \sum_d \sum_c \sum_s \sum_z \psi^{(d)} \pi^{(c)} \theta^{(c,s)} \phi_{w_1}^{(s,z)} \phi_{w_2}^{(s,z)}
\end{align}
%


\subsection{Inference}

We use collapsed Gibbs sampling \cite{griffiths2004finding} to estimate the latent variables $\theta$, $\phi$, $\pi$, and $\psi$. We first sample the clusters of the documents. At each transition step of the Markov chain, the cluster $c$ of the $d$th document is chosen according to the conditional probability shown in Equation \ref{eq:cluster}, where $l$ is the currently sampled cluster. The variable $N_{ab}^{AB}$ represents the number of $a \in A$ and $b \in B$ assignments and the variable $N^{AB}$ represents the sum of the number of $a \in A$ and $b \in B$ assignments for all possible $a$ and $b$, excluding the current $d$th document.
\begin{multline}
    \label{eq:cluster}
    P(c_d=l|rest) \approx \frac{N_l^C}{N^C + C\delta}
    \frac{\prod_s^S \prod_{x=1}^{N_{ds}^{DS}} \big(N_{ls}^{CS} + \gamma + x - 1\big)}{\prod_{y=0}^{N^{DS}} \big(N^{CS} + S\gamma + y - 1\big)} \\
    \frac{\prod_s^S \prod_z^T \prod_{x=1}^{N_{dsz}^{DST}} \big(N_{lsz}^{CST} + \theta + x - 1\big)}{\prod_{y=0}^{N^{DST}} \big(N^{CST} + T\theta + y - 1\big)}
\end{multline}
We then sample the aspect and sentiment of the word pairs. At each transition step of the Markov chain, the sentiment $s$ and the aspect $z$ of the $i$th word pair are chosen according to the conditional probability shown in Equation \ref{eq:sentasp}, where $j$ and $k$ are the currently sampled sentiment and aspect, respectively. The variables $N_{ab}^{AB}$ and $N^{AB}$ are similar to the description above, but now excluding the current $i$th word pair.
\begin{multline}
    \label{eq:sentasp}
    P(s_i=j,z_i=k|rest) \approx
    \frac{N_{cjk}^{CST} + \alpha} {N^{CST} + T\alpha}
    \frac{N_{cj}^{CS} + \gamma} {N^{CS} + S\gamma}\\
    \frac{N_{jkw_1}^{STW} + \beta_{sw_1}} {N^{STW} + \beta_{s}}
    \frac{N_{jkw_2}^{STW} + \beta_{sw_2}} {N^{STW} + \beta_{s} + 1}
\end{multline}
After inference is done, the approximate probability of sentiment $s$ in document $d$ is shown in Equation \ref{eq:piprob}. The approximate probability of aspect $z$ for sentiment $s$ in document $d$ is shown in Equation \ref{eq:thetaprob}. Finally, the approximate probability of a word $w$ in sentiment $s$ and aspect $z$ is shown in Equation \ref{eq:phiprob}.
\begin{equation}
    \label{eq:piprob}
    P(s|d) = \frac{N_{ds}^{DS} + \gamma} {N^{DS} + S\gamma}
\end{equation}
\begin{equation}
    \label{eq:thetaprob}
    P(z|s,d) = \frac{N_{dsz}^{DST} + \alpha} {N^{DST} + T\alpha}
\end{equation}
\begin{equation}
    \label{eq:phiprob}
    P(w|s,z) = \frac{N_{szw}^{STW} + \beta_{sw}} {N^{STW} + \beta_{s}}
\end{equation}
\section{Experimental setup}

In our experiments, we address the following research questions:

\begin{itemize}
    \item Does MicroASM perform well on aspect-level ABSA tasks such as aspect term extraction? (Section \ref{sec:aspectexp})
    \item Does MicroASM perform well on document-level ABSA tasks such as sentiment classification? (Section \ref{sec:documentexp})
\end{itemize}

\subsection{Data}

We use two kinds of datasets: Yelp\footnote{\url{https://www.yelp.com/dataset_challenge}} and Naver\footnote{\url{http://movie.naver.com/}} reviews dataset. The Yelp dataset consists of English reviews on restaurants and shops. Since the Yelp dataset also contains longer reviews, we filter out reviews that have more than 140 characters. We choose three categories based on the amount of data available and diversity: Nightlife, Restaurant, and Shopping. The Naver dataset consists of Korean movie reviews. Reviews in this dataset are naturally short, since the website limits the users to write only 140 characters. We choose three genres based on the amount of data available and diversity: Crime, Comedy, and Fantasy. Statistics of the data is shown in Table \ref{tab:datastat}. We do basic preprocessing to the data by lemmatizing and retaining only nouns, verbs, adjectives, and adverbs. Stanford CoreNLP\footnote{\url{https://stanfordnlp.github.io/CoreNLP/}} and Komoran\footnote{\url{http://www.shineware.co.kr/?page_id=835}} are used as preprocessing tools for English and Korean texts, respectively. We also handle negations by adding a negating prefix (e.g. \textit{not\_}) in the neighbors (set to a window$=5$) of a negating word. We also get the review ratings of each dataset for testing. We assume that reviews with less than 5 (or 3 for Yelp) stars are negative reviews, and positive otherwise. Note that we do not use this information during the training of the model.

\begin{table}[t]
    \small
    \centering
    \begin{subtable}{0.5\textwidth}
        \centering
        \begin{tabularx}{\textwidth}{|l|YYYY|}
            \hline
            Category & \#Review & +/- Ratio & Ave. Len. & \#Word \\ \hline
            Nightlife & 24,512 & 5.63 & 12.87 & 54,750 \\ 
            Restaurants & 53,407 & 5.90 & 12.92 & 234,100 \\ 
            Shopping & 15,615 & 5.45 & 12.38 & 21,076 \\ \hline
        \end{tabularx}
        \caption{Yelp Restaurants and Shopping datasets}
    \end{subtable}
    \begin{subtable}{0.5\textwidth}
        \centering
        \begin{tabularx}{\textwidth}{|l|YYYY|}
            \hline
            Category & \#Review & +/- Ratio & Ave. Len. & \#Word \\ \hline
            Comedy & 6,901 & 7.45 & 7.63 & 35,685 \\ 
            Crime & 11,342 & 4.78 & 8.04 & 99,246 \\ 
            Fantasy & 8,932 & 6.45 & 7.95 & 47,767 \\ \hline
        \end{tabularx}
        \caption{Naver Movie datasets}
    \end{subtable}
    \caption{Statistics of the datasets. \textbf{\#Review:} the total number of reviews/documents within a category. \textbf{+/- Ratio}: the ratio between the number of positive and negative reviews. \textbf{Ave. Len.:} average length in words. \textbf{\#Word:} The total number of unique words within a category.}
    \label{tab:datastat}    
\end{table}

As an aspect sentiment model, our models need a sentiment lexicon to propagate the inference of the sentiment variable. In our model, we use \texttt{Paradigm+} which consists of the sentiment oriental paradigm words from the original \texttt{Paradigm} \cite{turney2003measuring} such as \textit{good} and \textit{bad}, and additional general affective and evaluative words such as \textit{love} and \textit{hate} \cite{jo2011aspect}. For our Korean dataset, we use a Korean-translated \texttt{Paradigm+}. The English and Korean \texttt{Paradigm+} list is listed in Table \ref{tab:paradigm}. We also include the negated seed words (i.e. words with \textit{not\_} prefix) to the opposite sentiment seed list. For example, we include the seed word \textit{not\_good} in the negative seed list. For domain-specific environments where the sentiment lexicon is different, existing work on domain-specific lexicon induction \cite{hamilton2016inducing} can be adopted.

\begin{CJK}{UTF8}{}
\CJKfamily{mj}
\begin{table*}[t]
    \small
    \centering
    \begin{tabularx}{\textwidth}{|l|l|X|}
    \hline
    \multirow{2}{*}{English} & + & \textbf{good, nice, excellent, positive, fortunate, correct, superior}, amazing, attractive, awesome, best, comfortable, enjoy, fantastic, favorite, fun, glad, great, happy, impressive, love, perfect, recommend, satisfied, thank, worth  \\ \cline{2-3}
                             & - & \textbf{bad, nasty, poor, negative, unfortunate, wrong, inferior}, annoy, complain, disappointed, hate, junk, mess, dislike, unworthy, problem, regret, sorry, terrible, trouble, unacceptable, upset, waste, worst, worthless  \\ \hline
    \multirow{2}{*}{Korean}  & + & 좋다 (good), 우수하다 (excellent), 긍정적 (positive), 운, 행운 (fortunate), 옳, 맞 (correct), 훌륭하다 (superior), 매력 (attractive), 대단하다 (awesome), 베스트, 짱, 최고 (best), 편안하다 (comfortable), 즐기다, 즐겁다, 재미, 재미있다 (enjoy), 기쁘다 (glad), 멋지다 (great), 행복, 행복하다 (happy), 인상적 (impressive), 사랑, 사랑하다 (love), 완벽하다 (perfect), 추천, 추천하다 (recommend), 만족, 만족하다, 만족스럽다 (satisfied), 감사하다, 고맙다 (thank), 가치, 보람 (worth) \\ \cline{2-3}
                             & - & 나쁘다 (bad), 더럽다 (nasty), 불쌍, 불쌍하다, 초라하다 (poor), 부정적 (negative), 불행, 불행하다 (unfortunate), 잘못되다, 잘못, 틀리다 (wrong), 괴롭다, 괴로움, 괴롭히다 (annoy), 불평하다, 불평 (complain), 실망, 실망하다, 실망스럽다 (disappointed), 싫다, 싫어하다, 밉다 (hate), 쓰레기 (junk), 똥, 혼란 (mess), 싫증 (dislike), 문제 (problem), 후회, 후회하다 (regret), 미안하다, 죄송, 죄송하다 (sorry), 불편 (trouble), 당황, 당황하다, 당황스럽다 (flustered), 낭비, 낭비되다 (waste), 최악 (worst), 가치없다 (worthless)  \\ \hline
    \end{tabularx}
    \caption{The original and the Korean-translated \texttt{Paradigm+} sentiment seed lists. The Korean words are accompanied with their corresponding English translations. Note that some of the English words are not translated into Korean because they are not directly translatable. The bold-faced words are the sentiment oriental paradigm words from the \texttt{Paradigm} list.}
    \label{tab:paradigm}
\end{table*}
\end{CJK}


\subsection{Parameter tuning}

We set the Dirichlet priors to the following values: $\alpha=0.1$, $\gamma=1$, and $\delta=0.1$, following \cite{jo2011aspect,zuo2016topic}. In the case of the $\beta$ prior, we set it to different values depending on the current sentiment and the current pair of words. Moreover, following \cite{jo2011aspect} for consistency, the following Dirichlet prior $\beta$ is used accordingly. If both words are non-sentiment words, $\beta=0.01$. If at least one word is a sentiment word, then if the sentiment matches the current sentiment, $\beta=0.1$. Otherwise, $\beta$ is set to zero. We set the number of clusters $C=500$, number of topics $T=15$, and number of sentiments $S=2$. We limit the pairing of words into a context window that we set to $5$. We then run the model with $1500$ iterations.

\subsection{Baselines}

For our baseline models, we use the following four widely used aspect-sentiment models:

\begin{itemize}
    \item \textbf{Joint Sentiment Topic (JST) Model} \cite{lin2009joint}: The most basic aspect sentiment model which extends the LDA topic model by adding a sentiment latent variable.
    \item \textbf{Aspect and Sentiment Unification Model (ASUM)} \cite{jo2011aspect}: The most well-known aspect sentiment model which extends the JST model by adding a sentence layer in order for one sentence to capture a single aspect and sentiment. However, since the number of sentences in a short review is usually limited to one, ASUM cannot be directly used in the experiments. We instead use short phrases of size $5$ to represent the sentences of ASUM.
    \item \textbf{Joint Aspect-based Sentiment Topic (JAST) Model} \cite{wang2016mining}: A fine-grained aspect-based sentiment topic model that separates the distribution of general sentiment terms, aspect terms and aspect-specific sentiment terms.
\end{itemize}

For fairness in comparison, all of the above use the same sentiment lexicons as used by MicroASM for both English and Korean datasets. Also, the same hyperparameter $\alpha$, $\gamma$, $\beta$ are used for the same purpose.

\section{Experiments}

\subsection{Aspect term extraction}
\label{sec:aspectexp}

Each word distribution $\phi$ constitutes to a specific aspect and a sentiment. These distributions should contain aspect terms related to the aspect of $\phi$, and aspect-specific sentiment terms related to the sentiment of $\phi$. We perform evaluations to check whether these terms match with their corresponding aspect and sentiment. We use two kinds of metrics: distribution sentiment accuracy and aspect category assignment metrics.

\paragraph{Distribution sentiment accuracy}

In order to check whether the aspect-specific sentiment terms correspond to the $\phi$ distribution's sentiment, we evaluate using the accuracy of the classification of $\phi$'s sentiment based on human judgement. Given the top five terms of a distribution, we let two annotators classify the distribution with either positive, negative, or undecidable, where annotators are asked to assign the class \textit{undecidable} to distributions unclassifiable as positive or negative. We then compare the human-annotated sentiments to the sentiment of the distribution, and calculate the accuracy. Since JAST produces three different $\phi$ distributions, we use the aspect-specific sentiment term $\phi$ distribution for comparison with other models.

The results are reported in Table \ref{tab:sentacc}. On all datasets, MicroASM outperforms all the other aspect sentiment models. Moreover, JST performs the worst among all the aspect sentiment models on all datasets. Interestingly, the overall results of all the models on the Naver datasets are comparatively lower than the results on the Yelp datasets.
Although we can optimize for better accuracy by providing better hand-crafted seed lists and better performing tokenizers instead of machine-translated seed lists and off-the-shelf Korean tokenizers, the goal of this evaluation is to show how straightforward it is to extend to other domains and languages: MicroASM still performs the best despite problems on the Korean dataset.

\begin{table}[t]
    \small
    \centering
    \begin{subtable}{0.5\textwidth}
        \centering
        \begin{tabular}{|l|cccc|}
            \hline
            Dataset & JST & ASUM & JAST & MicroASM \\
            \hline
            Nightlife & \textit{66.7} & 68.3 & 80.0 & \textbf{88.3} \\
            Restaurant & \textit{65.0} & 66.7 & 68.3 & \textbf{83.3} \\
            Shopping & \textit{61.7} & 73.3 & 65.0 & \textbf{86.7} \\
            Average & \textit{64.2} & 69.4 & 71.1 & \textbf{86.1} \\
            \hline
        \end{tabular}
        \caption{Yelp Restaurants and Shopping datasets}
    \end{subtable}
    \begin{subtable}{0.5\textwidth}
        \centering
        \begin{tabular}{|l|cccc|}
            \hline
            Dataset & JST & ASUM & JAST & MicroASM \\
            \hline
            Comedy & \textit{38.3} & 40.0 & 51.7 & \textbf{66.7} \\
            Crime & \textit{36.7} & 45.0 & 71.7 & \textbf{76.7} \\
            Fantasy & 40.0 & \textit{36.7} & 60.0 & \textbf{73.3} \\
            Average & \textit{38.3} & 40.6 & 61.1 & \textbf{72.2} \\
            \hline
        \end{tabular}
        \caption{Naver Movie datasets}
    \end{subtable}
    \caption{Distribution sentiment accuracy results}
    \label{tab:sentacc}
\end{table}

\paragraph{Aspect category assignment metrics}

We also verify whether the aspect terms are related to the aspect of the given $\phi$ distribution. We do this by measuring how diverse, specific, and agreeable the aspect categories judged by human annotators are. Given the top 20 terms of a distribution and a set of aspect categories which consists of specific and general aspects, we let two annotators classify the distribution with a given set of aspect categories. Only the Restaurant dataset is used because it is the only dataset with a publicly available hand-annotated set of aspect categories \cite{pontiki2016semeval}. We use the aspect term $\phi$ distribution of JAST for comparison in this experiment. We then compute the following metrics: diversity, specificity, and agreeability.

\textbf{Diversity} measures how different the annotated labels are. The measure gives a higher value if the number of distinct annotated labels is also high, considering the number of word distributions annotated as a specific label. We use the Shannon diversity index to measure the diversity. The measure is given as follows:

\begin{equation}
Diversity = -\sum_{a=1}^A{\frac{C_a}{N}\log\bigg(\frac{C_a}{N}\bigg)}
\end{equation}

where $A$ is the list of aspect labels, $C_a$ is the number of word distributions annotated as label $a$, and $N = T*S*2 = 60$ is the total number of word distributions.

\textbf{Specificity} measures how particular the annotated labels are. The measure gives a higher value if the number of distinct annotated labels are high, not considering the number of word distributions annotated as a specific label. We use a simple metric for the specificity measure as follows:

\begin{equation}
Specificity = \frac{N-N(general,other,none)}{N}
\end{equation}

where $N(general,other,none)$ is the number of annotations using the labels "general", "other", or "none".

\textbf{Agreeability}    measures the agreement between annotators with regards to their annotated labels. The measure gives a higher value if annotations of annotators are the same. We use Cohen's kappa coefficient for multiple categories \cite{kraemer1980}, defined as follows:

\begin{equation}
Agreeability = \frac{\bar{P}-P_e}{1-P_e} \frac{1-\bar{P}}{N * (1-P_e)}
\end{equation}

where $\bar{P}$ is the relative observed agreement among annotators and $P_e$ is the hypothetical probability of chance agreement.

The results are reported in Table \ref{tab:aspect}. Overall, MicroASM significantly outperforms all the other models. The results also show that JAST performs the worst among the models. This is especially very apparent on the agreeability metric, which means that the $\phi$ distributions extracted by JAST is hard to interpret. This contradicts the results presented by the original paper \cite{wang2016mining}. Moreover, although JAST is generally better than JST and ASUM on other tasks such as distribution sentiment accuracy as presented above, and sentiment classification as presented in Section \ref{sec:documentexp}, it performs the worst in this experiment. This may be because the model needs a larger sentiment seed list to perform well\footnote{In the JAST paper \cite{wang2016mining}, they used a larger opinion lexicon \cite{hu2004mining}.}, while other models only need a very small subset of sentiment words to work well. One may consider applying larger seed lists, which can be done with insignificant increase in time and space complexity, but with the following complications. First, a good seed list must only have aspect-independent sentiments such as \textit{good} and \textit{bad}. When starting with a larger list, it is inevitable that aspect-specific sentiments (e.g. \textit{fast} is positive in \textit{fast performance} but negative in \textit{fast battery consumption}) will be included in the list, with negative effects in the process. Second, a larger seed list may not be available in other low-resource languages such as Korean. Hence in this experiment, a smaller set of sentiment seeds is desirable for aspect sentiment models.

\begin{table}[t]
    \small
    \centering
    \begin{tabular}{|l|cccc|}
        \hline
        Metric & JST & ASUM & JAST & MicroASM \\
        \hline
        Diversity & 0.584 & 0.541 & \textit{0.436} & \textbf{0.702} \\
        Specificity & 0.583 & 0.650 & \textit{0.450} & \textbf{0.767} \\
        Agreeability & 0.618 & 0.669 & \textit{0.158} & \textbf{0.826} \\
        Average & 0.595 & 0.620 & \textit{0.348} & \textbf{0.765} \\
        \hline
    \end{tabular}
    \caption{Aspect category assignment metrics on Yelp Restaurant dataset}
    \label{tab:aspect}
\end{table}

\subsection{Sentiment classification}
\label{sec:documentexp}

In this section, we evaluate our models in terms of sentiment classification. To determine the sentiment $s$ of a review $d$, we can use Equation \ref{eq:piprob} to solve for $p(s|d)$. This will return a sentiment distribution consisting of two probabilities, one for positive sentiment and another for negative sentiment. We classify a review as positive if the positive sentiment probability is greater than the negative sentiment probability, and classify it as negative otherwise. We compare the performance of our models to JST, ASUM, and JAST.

The classification results are presented in Table \ref{tab:sentresults} in terms of accuracy. On all datasets, MicroASM performs the best. JST performs the worst on all datasets except for the Shopping dataset, where ASUM performs worse. One interesting finding is that ASUM, JAST, and MicroASM perform similarly on both English and Korean datasets, however JST performs worse on the Naver datasets. This may mean that on real world micro review datasets, JST performs worse than usual when it is used to normal datasets. Note that the main difference between the Yelp and the Naver datasets is that Naver reviews are forced to be written within a 140-character limit, while Yelp does not impose such constraint. This means that the Yelp dataset is synthetic and the Yelp short reviews are less natural than that of the Naver short reviews. Thus, the observations discussed above 
are more evident in Naver reviews than in Yelp reviews. On such datasets, MicroASM still performs better than other models, thus is a more appropriate aspect sentiment model for micro reviews.

\begin{table}[t]
    \small
    \centering
    \begin{subtable}{0.5\textwidth}
        \centering
        \begin{tabular}{|l|cccc|}
            \hline
            Dataset & JST & ASUM & JAST & MicroASM \\
            \hline
            Nightlife & \textit{70.6} & 73.4 & 76.3 & \textbf{81.9} \\
            Restaurant & \textit{69.1} & 69.3 & 72.4 & \textbf{81.3} \\
            Shopping & 68.4 & \textit{67.9} & 70.0 & \textbf{79.1} \\
            Average & \textit{69.4} & 70.2 & 72.9 & \textbf{80.8} \\
            \hline
        \end{tabular}
        \caption{Yelp Restaurants and Shopping datasets}
        \label{tab:sentyelp}
    \end{subtable}
    \begin{subtable}{0.5\textwidth}
        \centering
        \begin{tabular}{|l|cccc|}
            \hline
            Dataset & JST & ASUM & JAST & MicroASM \\
            \hline
            Comedy & \textit{57.6} & 71.0 & 80.2 & \textbf{87.2} \\ 
            Crime & \textit{57.2} & 70.7 & 70.3 & \textbf{77.0} \\ 
            Fantasy & \textit{51.5} & 73.5 & 74.9 & \textbf{78.8} \\ 
            Average & \textit{55.4} & 71.7 & 75.1 & \textbf{81.0} \\
            \hline
        \end{tabular}
        \caption{Naver Movie datasets}
        \label{tab:sentnaver}
    \end{subtable}
    \caption{Sentiment classification accuracy results}
    \label{tab:sentresults}
\end{table}

\section{Related work}

Latent Dirichlet allocation (LDA) \cite{blei2003latent} has been extended to answer problems in ABSA. One such extension is the Joint Sentiment Topic (JST) model \cite{lin2009joint} by adding a new sentiment latent variable, aside from the topic latent variable. 
An extension to JST is Aspect and Sentiment Unification Model (ASUM) \cite{jo2011aspect}. ASUM extends JST simply by adding a sentence layer, based on the assumption that grouping words into a sentence during inference provide much better aspect discovery. JST has also been extended by separately generating sentiment and aspect terms \cite{wang2015sentiment2}. Recently, ASUM has been extended to further improve its performance by modeling the transition between two sentences using Markov chains \cite{rahman2016hidden} and by separating the variables of general and specific aspect and sentiment words \cite{wang2016mining}, among others. 

LDA has also been extended to accommodate short texts. 
Past works involve aggregating tweets through Twitter metadata information such as user information \cite{hong2010empirical} and through calculated scores based on Twitter trends such as burst score \cite{mehrotra2013improving}. 
A more domain-independent approach is used in Biterm Topic Model (BTM) \cite{yan2013biterm}. BTM removes the document layer, and instead used biterms, unordered pairs of co-occurring words, to learn topics from the documents. Another way to approach the problem is to use pseudodocuments. This approach is used in Pseudodocument Topic Model (PTM) \cite{zuo2016topic}. Using pseudodocuments, short texts are implicitly aggregated to avoid the data sparsity problem. 

There are several aspect sentiment models for datasets with additional metadata information such as Twitter. One such model is Twitter Opinion Topic Model (TOTM) \cite{lim2014twitter}, where they incorporate hashtags to extract product opinions from Twitter data. Microblog Sentiment Topic Model (MSTM) specially accounts for tokens such as hashtags to model both sentiment and topic word distribution \cite{ahuja2016microblog}. A more problem-specific model used timeslices as additional variables to detect sentiment-aware topics from social media \cite{kaminka2016joint}. 

\section{Conclusion}

In this paper, we proposed Micro Aspect Sentiment Model (MicroASM), an aspect sentiment model specifically for short reviews. MicroASM assumes that sentiment-aspect word pairs are the building blocks of information in short reviews and that reviews can be grouped into one information sharing cluster. We showed that our model performed better than previous generative models on two kinds of Aspect-based Sentiment Analysis (ABSA) tasks: aspect term extraction and sentiment classification. We confirmed that MicroASM extracts more meaningful aspect-specific sentiment terms and aspect terms. We showed two further applications that can be done using MicroASM: adjective-noun pair-based review summarization and preference-based review clustering. MicroASM is capable of discovering common and rare adjective-noun pairs that can be used for summaries. Moreover, we also showed that the model is able to automatically cluster the documents based on their preferences. Finally, we reported the results of sub-models, PairASM and ClusterASM created based on the individual observations, on the three ABSA tasks and showed that both observations are necessary to perform well on both aspect-level tasks such as aspect term extraction and document-level tasks such as sentiment classification.


\section*{Acknowledgments}

This work was supported by Samsung Research Funding Center of Samsung Electronics under Project Number SRFC-IT1701-01.

\bibliography{icdm}
\bibliographystyle{IEEEtran}

\end{document}